# Comparative Study of Large Language Models on Chinese Film Script Continuation: An Empirical Analysis Based on GPT-5.2 and Qwen-Max

Yuxuan Cao[1]*    Zida Yang[2]*    Ye Wang[3]†

[1]School of Arts, Peking University    [2]Guanghua School of Management, Peking University    [3]School of Journalism and Communication, Wuhan University

*Equal contribution. †Corresponding author: wangye@whu.edu.cn

## Abstract

As large language models (LLMs) are increasingly applied to creative writing, their performance on culturally specific narrative tasks warrants systematic investigation. This study constructs the first Chinese film script continuation benchmark comprising 53 classic films, and designs a multi-dimensional evaluation framework comparing GPT-5.2 and Qwen-Max-Latest. Using a "first half → continue second half" paradigm with 3 samples per film, we obtained 303 valid samples (GPT-5.2: 157, 98.7% validity; Qwen-Max: 146, 91.8% validity). Evaluation integrates ROUGE-L, Structural Similarity, and LLM-as-Judge scoring (DeepSeek-Reasoner).

Statistical analysis of 144 paired samples reveals: Qwen-Max achieves marginally higher ROUGE-L (0.2230 vs 0.2114, d=-0.43); however, GPT-5.2 significantly outperforms in structural preservation (0.93 vs 0.75, d=0.46), overall quality (44.79 vs 25.72, d=1.04), and composite scores (0.50 vs 0.39, d=0.84). The overall quality effect size reaches large effect level (d>0.8).

GPT-5.2 excels in character consistency, tone-style matching, and format preservation, while Qwen-Max shows deficiencies in generation stability. This study provides a reproducible framework for LLM evaluation in Chinese creative writing.

# 1 Introduction

## 1.1 Research Background

The rapid advancement of Large Language Models (LLMs) is fundamentally transforming creative industries' production methods. From literary creation to film script development, LLMs have demonstrated powerful text generation capabilities (Brown et al., 2020; OpenAI, 2023). However, existing research predominantly focuses on creative writing tasks in English contexts, leaving a systematic empirical gap in Chinese-language tasks, particularly those involving unique cultural contexts such as Chinese film script generation.

Film scripts represent a highly structured text form with distinctive format conventions (scene headings, dialogue markers, stage directions) and narrative logic (character development, conflict progression, plot advancement). Chinese film scripts, while inheriting Hollywood's classical narrative paradigms, incorporate rich indigenous cultural elements and unique aesthetic pursuits (Dai, 2018). This cultural specificity makes Chinese film script continuation an ideal testbed for examining LLMs' cross-cultural creative generation capabilities.

## 1.2 Research Questions

This study addresses the following core questions:

**RQ1**: How do current mainstream LLMs perform on Chinese film script continuation tasks? Are there significant differences between models?

**RQ2**: Can LLMs effectively preserve the original work's format conventions and narrative style during script continuation?

**RQ3**: What are the key factors affecting model continuation quality? What is the consistency between automated evaluation metrics and LLM judgments?

## 1.3 Research Contributions

The main contributions of this study include:

1. **Dataset Construction**: Establishing the first standardized Chinese film script continuation benchmark dataset, covering 53 classic Chinese films across genres and eras, with automated Format Profile and Format Contract mechanisms.
2. **Evaluation Framework Innovation**: Proposing a multi-dimensional evaluation system integrating automated metrics (ROUGE-L, Structural Similarity) with LLM-as-Judge scoring, and introducing Entman's framing analysis theory for difference attribution, achieving a complete analytical pipeline from quantitative scoring to qualitative interpretation.
3. **Empirical Findings**: Based on rigorous paired experimental design and comprehensive statistical testing (including effect sizes and confidence intervals), this study first reveals performance differences between GPT-5.2 and Qwen-Max on Chinese film script continuation, providing empirical guidance for model selection in Chinese creative writing.
4. **Methodological Contribution**: Proposing a Chunked Generation strategy to address long-text generation challenges, and developing reproducible experimental procedures for future research reference.

## 1.4 Paper Structure

The remainder of this paper is organized as follows: Section 2 reviews related work; Section 3 details the research methodology, including dataset construction, model configuration, and evaluation metric design; Section 4 reports experimental setup and results; Section 5 discusses findings and their theoretical implications; Section 6 concludes and outlines future research directions.

## 2 Related Work

### 2.1 Large Language Models and Creative Writing

The application of LLMs in creative writing has become a research hotspot in natural language processing. The GPT series (Radford et al., 2019; Brown et al., 2020; OpenAI, 2023, 2025) has demonstrated powerful text generation capabilities, producing fluent and creative text in poetry, fiction, and script generation tasks. The Qwen series (Bai et al., 2023), as a representative Chinese-native LLM, has shown competitiveness in Chinese text understanding and generation tasks.

In story generation, Fan et al. (2018) proposed the Reddit WritingPrompts-based story generation task, providing an important benchmark for long-form narrative generation. Rashkin et al. (2020)'s STORIUM dataset further advanced collaborative story creation research. However, these works primarily target English texts, with Chinese story generation research remaining relatively scarce.

### 2.2 Script Generation and Format Preservation

Script generation is among the most challenging creative writing tasks, requiring simultaneous consideration of narrative logic, character development, and format conventions. Zhu et al. (2023) explored using LLMs for film script writing assistance, noting difficulties in maintaining long-range consistency. Mirowski et al. (2023) proposed the DRAMATRON system, improving script structural integrity through hierarchical generation strategies.

Format preservation poses a unique challenge in script generation. Unlike general story text, scripts have strict format conventions including scene headings, dialogue markers, and stage directions. Existing research has largely overlooked model performance at the format level—this study addresses this gap by designing the Structural Similarity metric.

### 2.3 Text Generation Evaluation Methods

Automated evaluation of text generation remains a persistent NLP challenge. ROUGE (Lin, 2004) and BLEU (Papineni et al., 2002), based on n-gram overlap, are widely used but their correlation with human judgment has been questioned (Liu et al., 2016). BERTScore (Zhang et al., 2020) uses pre-trained language model semantic representations for similarity computation, partially alleviating lexical-level metric limitations.

Recently, the LLM-as-Judge paradigm has gained widespread attention (Zheng et al., 2023). Using powerful models like GPT-4 as judges enables more fine-grained evaluation closer to human judgment. Chiang & Lee (2023) demonstrated that with appropriate prompt engineering, LLM judgments achieve high consistency with human expert evaluations. This study employs DeepSeek-Reasoner as the judge model, combined with Entman's framework theory for structured evaluation.

### 2.4 Entman's Framing Theory

Entman (1993)'s framing theory has been widely applied in communication research, with four core elements: Problem Definition, Causal Interpretation, Moral Evaluation, and Treatment Recommendation. This study introduces Entman's framework into qualitative analysis of LLM-generated text, explaining sources of differences between model continuations and original works.

## 2.5 Chinese NLP and Cultural Specificity

Chinese natural language processing presents unique challenges, including word segmentation, disambiguation, and cultural background understanding (Qiu et al., 2020). In creative writing, cultural specificity is particularly pronounced. Chinese film scripts involve not only linguistic expression but also historical allusions, social customs, and value systems. Li et al. (2024) demonstrated that even the most advanced multilingual models exhibit understanding biases when processing Chinese-specific cultural concepts.

# 3 Methodology

## 3.1 Dataset Construction

### 3.1.1 Data Source and Preprocessing

This study collected Chinese film script raw texts from public sources, spanning different periods and genres from 1922 to 2021. The preprocessing pipeline includes:

1. **Text Cleaning**: Unified encoding format (UTF-8), removal of irrelevant annotations and format noise
2. **Upper-Lower Split**: Splitting each script at content midpoint into "first half" (U) and "second half" (D)
3. **Quality Filtering**: Retaining scripts with complete upper and lower segments

The final dataset comprises 53 valid film scripts covering comedy, action, romance, martial arts, historical, and other genres. Table 1 presents basic dataset statistics.

**Table 1. Dataset Statistics**

| Statistic | Value |
| --- | --- |
| Total Films | 53 |
| Mean First-Half Characters | 19,835 |
| Mean Second-Half Characters | 14,599 |
| Longest Script (characters) | 71,875 (*Detective Chinatown in India*) |
| Shortest Script (characters) | 12,761 (*Happy Together*) |
| Year Range | 1922-2021 |

### 3.1.2 Format Profile and Format Contract

To achieve automated detection and transmission of format conventions, this study designs two core mechanisms:

**Format Profile**: Automatically detects and records format features of each script, including:

- Scene heading style (e.g., "**1**, Scene", "1. Scene, Day, Interior", "Scene X")
- Dialogue marker method ("Character: Dialogue" or "Character (newline) Dialogue")
- Stage direction symbols (△, ▲, or parentheses)

- Blank line policy and emphasis symbols

**Format Contract**: Converts the Format Profile into natural language description, embedded in generation prompts to guide models in following format conventions.

The Format Profile detection algorithm, based on regular expressions and heuristic rules, identifies five major scene heading styles, three dialogue marker methods, and three stage direction types, forming a complete format feature vector.

### 3.1.3 Data Index Structure

All preprocessed data is stored in JSONL format, with each record containing:

- Film ID and title
- Upper/lower segment file paths and character statistics
- Format Profile (JSON object)
- Format Contract path

## 3.2 Model Configuration

### 3.2.1 Experimental Models

This study selects two representative LLMs for comparison:

**GPT-5.2** (OpenAI, 2025): OpenAI's latest flagship model released in December 2025, excelling in professional knowledge work and complex reasoning tasks. This study uses the Thinking mode version via API, which features enhanced deep reasoning capabilities.

**Qwen-Max-Latest** (Alibaba Cloud, 2025): The flagship model of the Tongyi Qianwen series, representing Chinese-native LLMs with certain advantages in Chinese understanding and generation tasks.

Both models are called through a unified API endpoint (api.shubiaobiao.cn) to ensure experimental environment consistency.

### 3.2.2 Prompt Design

To ensure fair comparison, this study designs **identical** prompt templates for GPT-5.2 and Qwen-Max, including:

**System Prompt**:

```
You are a senior film screenwriter and script formatting editor. Your sole task is: strictly
following the Format Contract of the input script "Part I", directly continue the subsequent
plot "Part II". You must output JSON conforming to requirements and provide no additional
explanation.
```

**User Prompt** contains five hard constraints:

1. Format hard constraint: Follow Format Contract item by item
2. Continuity hard constraint: Continue character settings, relationships, unresolved conflicts
3. Narrative hard constraint: Maintain genre atmosphere and linguistic style
4. Boundary hard constraint: No meta-discourse

5. Length constraint: Continuation length 60%-90% of first half

### 3.2.3 Chunked Generation Strategy

Due to API max_tokens limitations (Qwen-Max ~8000 tokens), this study employs a Chunked Generation strategy:

1. Calculate target length range: $[0.6 \times L_{\text{up}}, 0.9 \times L_{\text{up}}]$
2. Each segment request generates 3500-6500 characters
3. Pass the last 4000 characters of generated content as context
4. Execute maximum 10 chunked calls
5. Stop when cumulative length reaches target minimum

## 3.3 Evaluation Metric Design

### 3.3.1 Automated Metrics

**ROUGE-L**: F1 score based on Longest Common Subsequence, calculated after jieba tokenization. This metric measures lexical overlap between continuation text and actual second half.

**Structural Similarity**: A novel metric proposed in this study, extracting five structural features based on Format Profile:

- Scene heading ratio (scene_ratio)
- Dialogue line ratio (dialogue_ratio)
- Blank line ratio (blank_ratio)
- Stage direction ratio (stage_ratio)
- Bold symbol density (bold_density)

Structural Similarity formula:

$$S_{\text{struct}} = 1 - \frac{\sum_k |f_k^{\text{gen}} - f_k^{\text{ref}}|}{\sum_k |f_k^{\text{ref}}| + \epsilon}$$

where $f_k$ represents the $k$-th feature value, and $\epsilon = 10^{-6}$ is a smoothing term. This metric ranges from $[0, 1]$, with higher values indicating greater structural similarity.

### 3.3.2 LLM-as-Judge Evaluation

This study employs DeepSeek-Reasoner as the judge model, known for its deep reasoning capabilities, suitable for complex text quality evaluation. Judge output includes structured JSON scores:

**Primary Metric**:

- Overall similarity (overall_similarity_0_100)

**Sub-dimension Scores** (each 0-100):

- Plot event alignment (plot_event_alignment)
- Character consistency (character_consistency)
- Tone-style match (tone_style_match)

- Format match (format_match)
- Ending closure (ending_closure)

**Difference Evidence** (diff_evidence): Cited short segments from Reference and Generated

**Mechanism Attribution** (mechanism_attribution): Four-element analysis based on Entman's framework

### 3.3.3 Composite Score Calculation

To integrate multi-dimensional evaluation results, this study defines a composite score:

$$S_{\text{composite}} = 0.4 \times \text{ROUGE}_L + 0.3 \times S_{\text{struct}} + 0.3 \times \frac{S_{\text{overall}}}{100}$$

Weight settings are based on: ROUGE-L as an objective lexical-level metric receives larger weight; Structural Similarity and Overall Quality Score each receive 30%, balancing format preservation and content quality.

## 3.4 Experimental Design

### 3.4.1 Sample Generation

For 53 films, each model generates 3 independent samples (temperature=0.7 to ensure creative diversity), theoretically yielding 318 samples. Validity criteria:

- Continuation length within target range (60%-90%)
- No meta-discourse (e.g., "Here is the continuation")
- Successful JSON parsing
- Successful API call (no timeout or content moderation blocking)

### 3.4.2 Paired Comparison Design

To control for film-specific factors (genre, style, length, etc.) as potential confounds, this study employs a paired comparison design: statistical tests are conducted only on samples where both models successfully generated output for the same film and sample index. This design ensures comparison fairness and conclusion reliability.

### 3.4.3 Statistical Methods

- Descriptive statistics: Mean, Standard Deviation
- Paired t-test: Testing significance of between-model differences
- Effect size calculation: Cohen's d for practical difference magnitude
  - |d| < 0.2: Negligible effect
  - 0.2 ≤ |d| < 0.5: Small effect
  - 0.5 ≤ |d| < 0.8: Medium effect
  - |d| ≥ 0.8: Large effect
- 95% Confidence Intervals: Assessing precise difference ranges

# 4 Experimental Results

## 4.1 Data Collection Overview

Table 2 presents the final data collection results.

**Table 2. Sample Collection Statistics**

| Metric | GPT-5.2 | Qwen-Max | Total |
| --- | --- | --- | --- |
| Theoretical Samples | 159 | 159 | 318 |
| Valid Samples | 157 | 146 | 303 |
| Validity Rate | 98.7% | 91.8% | 95.3% |
| Paired Samples | 144 | 144 | 144 pairs |

GPT-5.2 demonstrates higher generation stability, with only 2 samples marked invalid due to insufficient length (validity rate 98.7%); Qwen-Max has 13 invalid samples (validity rate 91.8%), primarily due to API timeout (HTTP 524 error, ~46%) and content moderation blocking (HTTP 400 error, ~31%). The validity rate difference between models is practically significant (Δ=6.9 percentage points).

## 4.2 Automated Metric Results

Table 3 presents both models' performance on automated metrics.

**Table 3. Automated Metric Comparison (Full Sample)**

| Metric | GPT-5.2 (n=157) | Qwen-Max (n=146) | Difference |
| --- | --- | --- | --- |
| ROUGE-L | 0.2127 ± 0.0505 | 0.2233 ± 0.0501 | -0.0106 |
| Structural Similarity | 0.9193 ± 0.1866 | 0.7508 ± 0.3495 | +0.1685 |

Full-sample statistics show Qwen-Max slightly higher on ROUGE-L, while GPT-5.2 has a clear advantage in Structural Similarity. Notably, Qwen-Max's Structural Similarity standard deviation (0.3495) far exceeds GPT-5.2's (0.1866), indicating high variability in format preservation performance.

## 4.3 LLM-as-Judge Scoring Results

Table 4 presents detailed scores from DeepSeek-Reasoner evaluation.

**Table 4. Judge Score Comparison (Full Sample)**

| Dimension | GPT-5.2 (n=157) | Qwen-Max (n=146) | Difference |
| --- | --- | --- | --- |
| Overall Score | 44.99 ± 14.41 | 25.81 ± 12.05 | +19.18 |
| Plot Alignment | 29.76 ± 17.67 | 13.18 ± 11.62 | +16.58 |
| Character Consistency | 44.59 ± 18.52 | 22.21 ± 14.81 | +22.38 |

| Dimension | GPT-5.2 (n=157) | Qwen-Max (n=146) | Difference |
|---|---|---|---|
| Tone-Style Match | 56.90 ± 16.41 | 30.21 ± 15.71 | +26.69 |
| Format Match | 80.69 ± 14.80 | 63.86 ± 22.50 | +16.83 |
| Ending Closure | 12.44 ± 15.83 | 6.08 ± 8.11 | +6.36 |

GPT-5.2 significantly outperforms Qwen-Max across all scoring dimensions. Figure 1 visually illustrates performance differences between models across dimensions using radar and bar charts.

**Figure 1. Multi-dimensional Score Comparison**

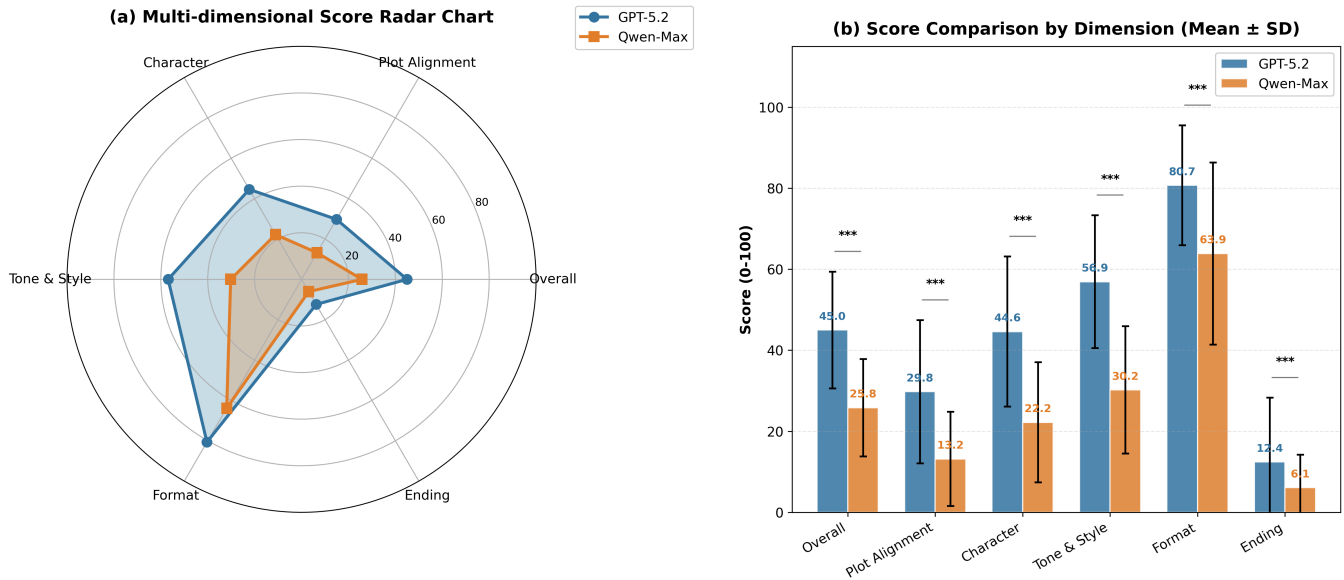

*Figure 1 Caption: (a) Radar chart showing overall profiles across six scoring dimensions; (b) Grouped bar chart showing mean ± SD for each dimension, \*\*\* indicates p<0.001.*

The largest difference occurs in **Tone-Style Match** (+26.69 points), indicating GPT-5.2 better captures the original work's narrative tone and linguistic style; followed by **Character Consistency** (+22.38 points), demonstrating GPT-5.2's superiority in maintaining character personality and motivation coherence.

Notably, both models score low on **Ending Closure** (GPT-5.2: 12.44; Qwen-Max: 6.08), reflecting current LLMs' shared challenge in long-range narrative conclusion.

## 4.4 Paired Statistical Tests

Table 5 presents statistical test results based on 144 paired samples, including complete effect sizes and confidence intervals.

**Table 5. Paired t-test Results (n=144 pairs)**

| Metric | GPT-5.2 | Qwen-Max | Difference [95% CI] | t-value | p-value | Cohen's d |
|---|---|---|---|---|---|---|
| ROUGE-L | 0.2114±0.0497 | 0.2230±0.0502 | -0.012 [-0.016, -0.007] | -5.12 | <0.001 | -0.43 (Small) |
| Structural Sim. | 0.9299±0.1562 | 0.7473±0.3495 | +0.183 [+0.118, +0.247] | 5.53 | <0.001 | +0.46 (Small) |

| Metric | GPT-5.2 | Qwen-Max | Difference [95% CI] | t-value | p-value | Cohen's d |
|---|---|---|---|---|---|---|
| Overall Score | 44.79±14.42 | 25.72±12.06 | +19.07 [+16.08, +22.06] | 12.45 | <0.001 | **+1.04 (Large)** |
| Composite Score | 0.4979±0.0702 | 0.3906±0.1156 | +0.107 [+0.087, +0.128] | 10.09 | <0.001 | **+0.84 (Large)** |

Note: Effect size interpretation: |d|<0.2 negligible, 0.2≤|d|<0.5 small, 0.5≤|d|<0.8 medium, |d|≥0.8 large

Figure 2 visualizes effect sizes and 95% confidence intervals for each metric using a forest plot.

**Figure 2. Effect Size Forest Plot**

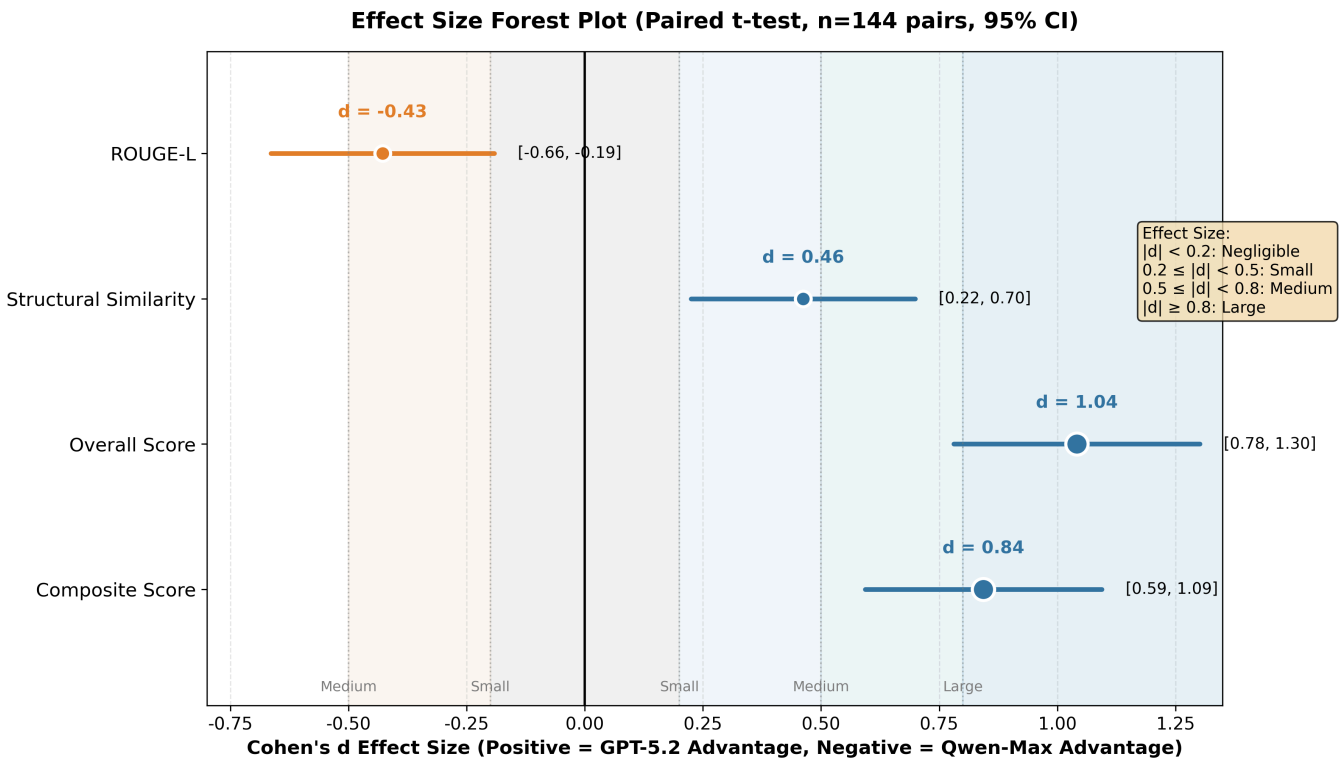

*Figure 2 Caption: Forest plot showing Cohen's d effect sizes and 95% confidence intervals for four core metrics. Positive values indicate GPT-5.2 advantage (blue), negative values indicate Qwen-Max advantage (orange). Background color regions mark effect size thresholds. Overall Score and Composite Score effect sizes reach large effect level (|d|≥0.8).*

**Key Findings**:

1. **All metric differences achieve high statistical significance (p<0.001)**.

2. **ROUGE-L is the only metric where Qwen-Max outperforms GPT-5.2**, but the effect size is small (d=-0.43), with a narrow confidence interval (difference ~0.7-1.6 percentage points), indicating limited practical impact.

3. **Overall Quality Score effect size reaches large effect level (d=1.04)**—this represents one of the study's most important findings, demonstrating GPT-5.2's content quality advantage has substantial practical significance.

4. **Composite Score effect size also reaches large effect level (d=0.84)**, with GPT-5.2's composite score approximately 27.5% higher than Qwen-Max (0.4979 vs 0.3906).

## 4.5 Composite Score Analysis

Figure 3 presents the composite score distribution comparison between models.

**Figure 3. Composite Score Distribution Comparison**

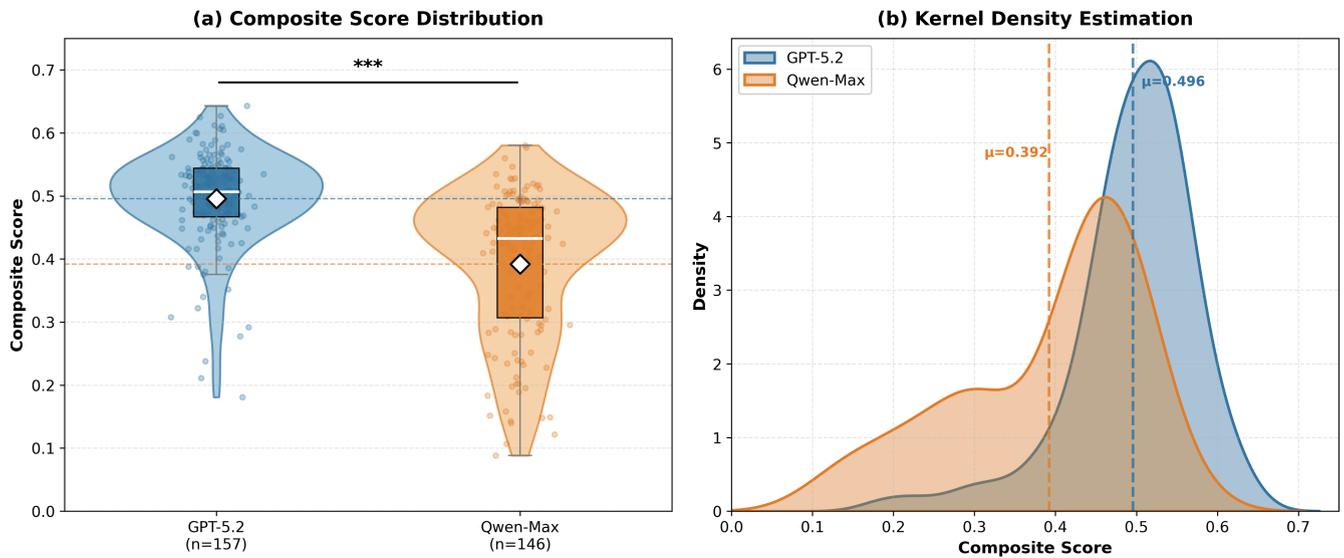

*Figure 3 Caption: (a) Violin + box plot showing composite score distributions, white diamonds indicate means, \*\*\* indicates p<0.001; (b) Kernel density estimation curves with dashed lines marking model means.*

**Table 6. Composite Score Descriptive Statistics**

| Statistic | GPT-5.2 | Qwen-Max |
| --- | --- | --- |
| Mean | 0.4958 | 0.3920 |
| Std Dev | 0.0763 | 0.1158 |
| Minimum | 0.18 | 0.09 |
| Maximum | 0.64 | 0.58 |
| IQR | 0.07 | 0.14 |
| CV | 15.4% | 29.5% |

GPT-5.2's composite score is not only higher on average but also more concentrated (SD 0.0763 vs 0.1158, CV 15.4% vs 29.5%). Qwen-Max's high variance (CV ~30%) indicates substantial performance fluctuation across different films, suggesting generation quality stability requires improvement.

## 4.6 Case Study Analysis

### Case 1: *Farewell My Concubine* (ID=02, Sample 1)

**Context**: Classic art film with melancholic narrative style, nuanced psychological portrayal, using "Character (newline) Dialogue" format.

| Metric | GPT-5.2 | Qwen-Max | Difference |
| --- | --- | --- | --- |
| ROUGE-L | 0.153 | 0.203 | -0.050 |
| Structural Similarity | 0.991 | 0.427 | **+0.564** |
| Overall Score | 72 | 25 | **+47** |

**Analysis**: GPT-5.2 successfully preserved the original's "Character (newline) Dialogue" format style (Structural Similarity 0.991) and continued the complex emotional entanglement between Cheng Dieyi and Duan Xiaolou; Qwen-Max showed significant format deviation (Structural Similarity only 0.427), with inconsistent dialogue markers and introduction of lighter elements inconsistent with the original tone. This case exemplifies format preservation's impact on overall quality.

#### Case 2: *Detective Chinatown* (ID=01, Sample 0)

**Context**: Comedy mystery film with fast pacing, dialogue-dense, using "Character: Dialogue" format.

| Metric | GPT-5.2 | Qwen-Max | Difference |
| --- | --- | --- | --- |
| ROUGE-L | 0.209 | 0.214 | -0.005 |
| Structural Similarity | 0.996 | 0.999 | -0.003 |
| Overall Score | 32 | 25 | +7 |

**Analysis**: Both models performed comparably on this film, both well-preserving the "Character: Dialogue" format (Structural Similarity both >0.99). Similar ROUGE-L indicates comparable lexical overlap, but GPT-5.2 maintains advantage in plot progression reasonability and character consistency, reflected in the Overall Score gap (+7 points). This case shows that when format conventions are relatively simple, the gap between models narrows.

# 5 Discussion

## 5.1 Main Findings and Theoretical Implications

### 5.1.1 Dissociation Between Lexical Overlap and Quality Judgment

This study reveals an intriguing dissociation between ROUGE-L and Overall Quality Score: Qwen-Max has a slight advantage on ROUGE-L (d=-0.43) but significantly trails on LLM-judged overall quality (d=+1.04). Figure 4 visualizes this dissociation phenomenon.

**Figure 4. ROUGE-L vs Overall Score Dissociation**

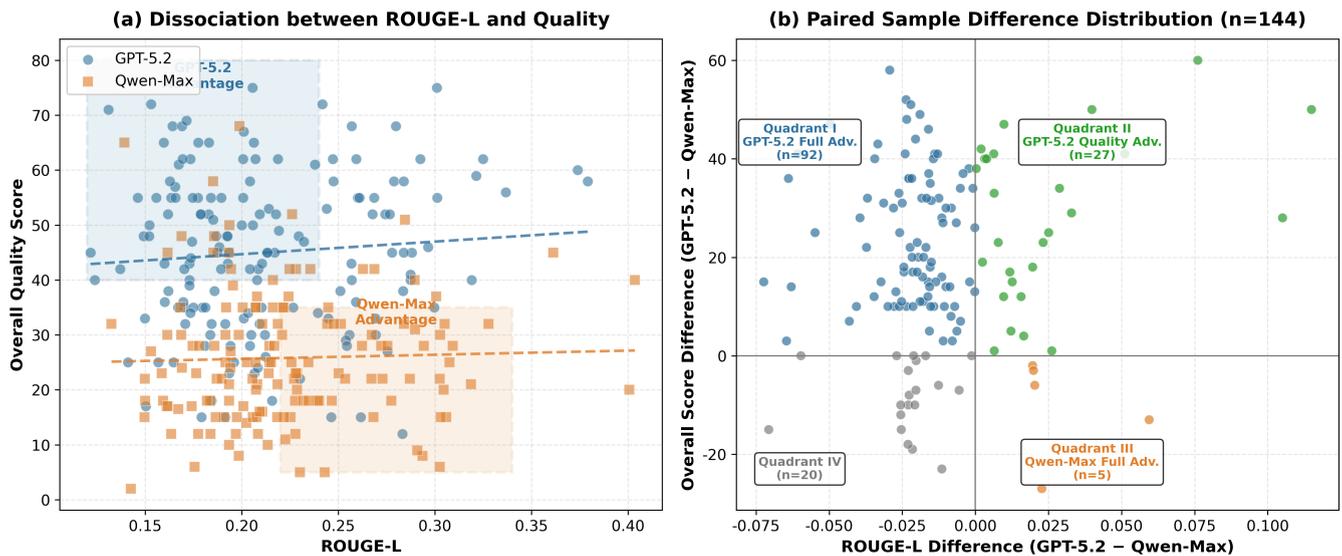

*Figure 4 Caption: (a) Scatter plot showing relationship between ROUGE-L and Overall Quality Score, with low correlations for both models (GPT-5.2: r=0.08; Qwen-Max: r=0.03), dashed lines show regression fits; (b) Paired sample difference distribution, x-axis shows ROUGE-L difference, y-axis shows Overall Score difference, four quadrants labeled with sample counts, majority falling in Quadrant I (GPT-5.2 full advantage).*

This finding corroborates Liu et al. (2016)'s discussion of n-gram metric limitations—high lexical overlap does not equate to high-quality continuation.

A possible explanation: Qwen-Max tends to "borrow" more original vocabulary and expressions, elevating ROUGE-L scores; but this borrowing may sacrifice narrative coherence and creativity. In contrast, GPT-5.2, while using more "new words," performs better at maintaining narrative logic and character consistency. This finding has important implications for text generation evaluation method design: **single lexical-level metrics are insufficient for comprehensive creative writing quality assessment**.

### 5.1.2 Significant Differences in Structural Preservation Capability

GPT-5.2's absolute advantage in Structural Similarity (0.93 vs 0.75) represents one of this study's most notable findings. Figure 5 shows detailed distribution and per-film comparison of Structural Similarity.

**Figure 5. Structural Similarity Comparison**

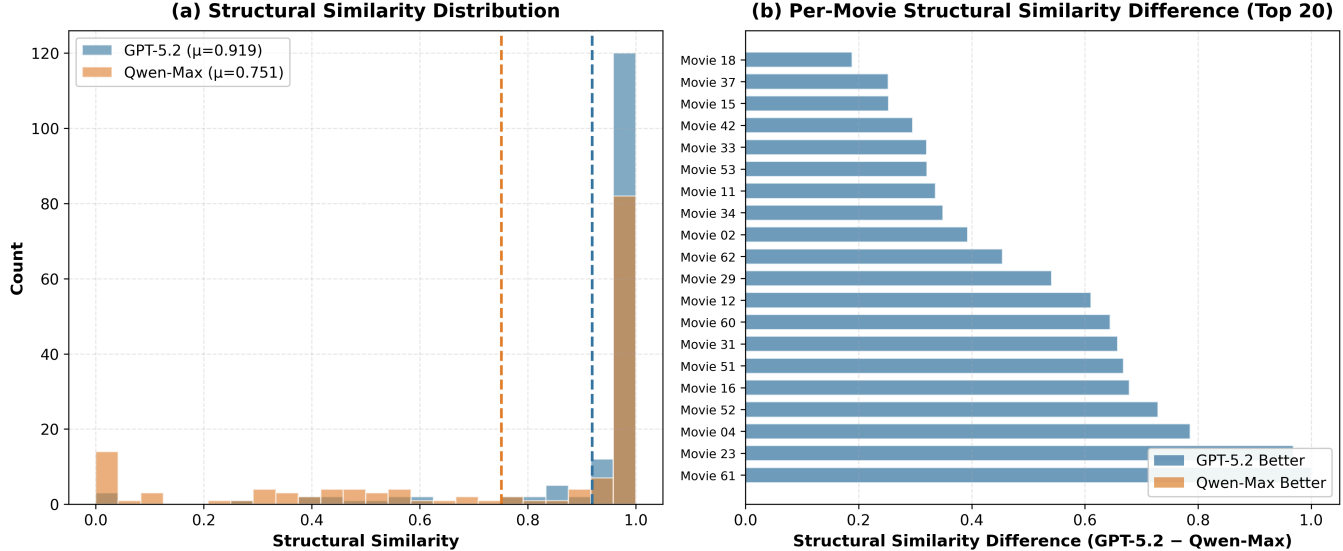

Figure 5 Caption: (a) Structural Similarity histogram distribution comparison, GPT-5.2 concentrated in high-score region (mean 0.919), Qwen-Max more dispersed (mean 0.751); (b) Paired differences by film ID (showing top 20 with largest differences), blue indicates GPT-5.2 better, orange indicates Qwen-Max better.

This indicates GPT-5.2 possesses stronger format convention perception and compliance capabilities, effectively extracting and reproducing format features from the first half text.

From a technical perspective, this may reflect GPT-5.2's advantage in **Instruction Following** capability. The Format Contract is embedded in prompts as natural language—models must understand and execute these format constraints, and GPT-5.2 demonstrates more robust performance. Additionally, Qwen-Max's high Structural Similarity variance (0.35 vs 0.19) indicates instability in format understanding.

### 5.1.3 Differences in Cultural Context Understanding

Although Qwen-Max, as a Chinese-native model, was expected to have advantages on Chinese tasks, experimental results do not support this hypothesis. Possible explanations include:

1. **Task Specificity**: Script continuation involves complex narrative reasoning and character development beyond pure language understanding, requiring stronger long-range dependency modeling.
2. **Model Scale and Training Data**: GPT-5.2's parameter scale and training data volume may compensate for language prior disadvantages, with more thorough training on complex reasoning tasks.
3. **Instruction Following Capability**: This task demands high format constraints and long-text coherence—GPT-5.2's Thinking mode advantages in deep reasoning manifest here.

## 5.2 Difference Attribution Based on Entman's Framework

Through analysis of mechanism attribution data from DeepSeek-Reasoner, we identify typical differences between models on Entman's four framework elements:

**Problem Definition**: GPT-5.2 tends to continue developing the original's core conflicts, identifying and extending unresolved dramatic tension; Qwen-Max sometimes introduces new problem threads unrelated to the original, causing narrative focus dispersion.

**Causal Interpretation**: GPT-5.2's causal chains are more coherent, with character behavior motivations consistent with first-half settings; Qwen-Max occasionally shows character motivation drift, with behavioral logic inconsistent with prior settings.

**Moral Evaluation**: GPT-5.2 better preserves the original's value stance and moral framework; Qwen-Max shows value framework shifts in some cases, possibly related to its content moderation mechanisms.

**Treatment Recommendation**: Both models perform poorly on narrative conclusion, with uniformly low "Ending Closure" scores (GPT-5.2: 12.44; Qwen-Max: 6.08), reflecting current LLMs' shared limitations in long-range narrative planning.

### 5.2.5 Correlation Analysis Among Evaluation Metrics

Figure 6 presents correlations among evaluation metrics and comprehensive performance comparison between models.

**Figure 6. Comprehensive Analysis Heatmap**

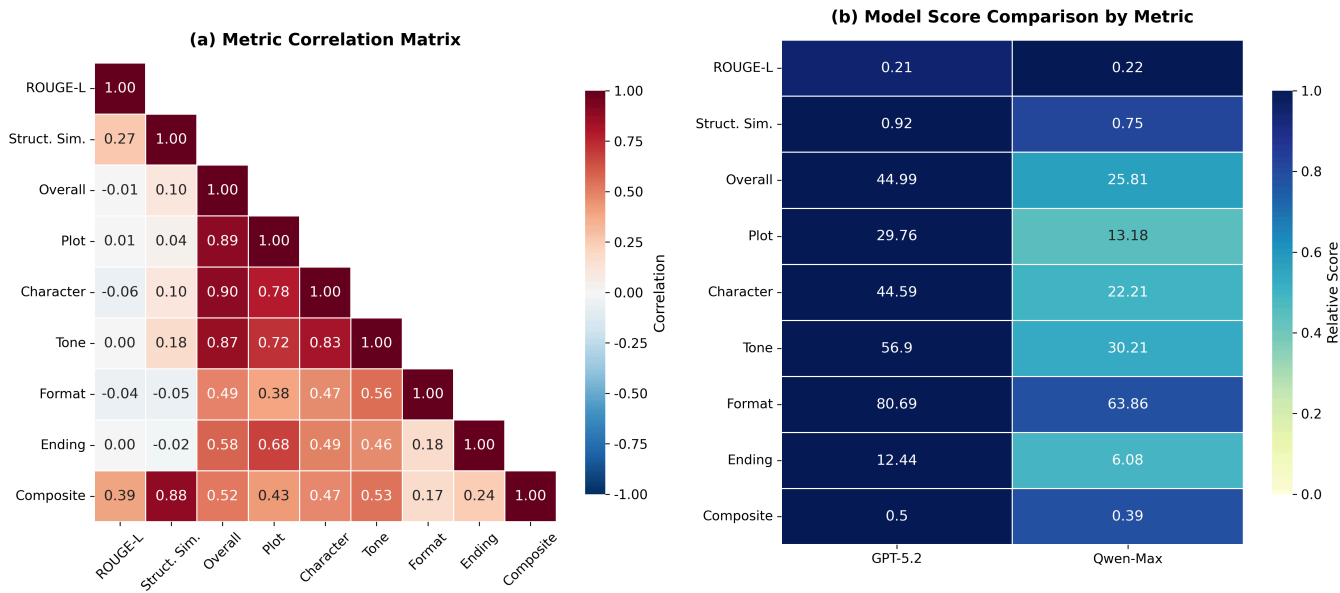

*Figure 6 Caption: (a) Evaluation metric correlation matrix (lower triangle), color intensity represents Pearson correlation coefficient, red for positive, blue for negative correlation; (b) Model score comparison heatmap by metric, values show actual means, color intensity indicates row-relative scores.*

The correlation matrix reveals that Overall Score correlates moderately to highly with sub-dimensions (Character Consistency, Tone-Style, Format Match), indicating good internal consistency of LLM-as-Judge scoring. Notably, ROUGE-L shows low correlation with other quality metrics (r<0.3), further confirming lexical overlap metrics' limitations in creative writing evaluation.

## 5.3 Limitations

This study has the following limitations:

1. **Judge Bias**: DeepSeek-Reasoner as the judge model may exhibit systematic biases (e.g., style preferences); future research could introduce human evaluation for cross-validation.
2. **Dataset Scale and Representativeness**: While 53 films cover multiple genres, the scale is limited (144 paired samples total), potentially affecting conclusion generalizability. Future work could expand dataset scale and conduct stratified analysis across eras and genres.
3. **Single Task Paradigm**: This study examines only "upper→lower" continuation, excluding other script generation scenarios (scene expansion, dialogue generation, script revision, etc.).
4. **Model Version Iteration**: LLMs update rapidly; this study's conclusions may change with model upgrades. Future research should establish continuous benchmarking mechanisms.
5. **Lack of Human Evaluation**: This study relies primarily on automated metrics and LLM judgment, lacking professional screenwriter evaluation, potentially affecting evaluation credibility.

## 5.4 Practical Implications

For creative industry practitioners, this study offers the following practical insights:

1. **Model Selection**: For script creation tasks requiring strict format conventions and long-text coherence, GPT-5.2 may be the better choice, particularly for professional scenarios requiring specific format style preservation.
2. **Evaluation Strategy**: ROUGE and other lexical-level metrics should not be solely relied upon for script

quality evaluation; structural analysis and content quality judgment should be combined to construct multi-dimensional evaluation systems.

3. **Hybrid Strategy**: Consider using Qwen-Max for initial draft generation (leveraging its ROUGE-L and cost advantages), then GPT-5.2 for format correction and quality enhancement.

4. **Human-AI Collaboration**: LLMs' universal deficiencies in Ending Closure indicate that human creator participation remains indispensable, especially for narrative conclusion and thematic elevation. LLMs are more suitable as creative assistance tools rather than complete replacements.

# 6 Conclusion and Future Work

## 6.1 Summary

This study constructs the first Chinese film script continuation benchmark dataset and systematically compares GPT-5.2 and Qwen-Max performance on this task. Based on empirical analysis of 303 valid samples and 144 paired samples, the main conclusions are:

1. **Overall Performance**: GPT-5.2 significantly outperforms Qwen-Max on composite score (0.4979 vs 0.3906, p<0.001, Cohen's d=0.84), with effect size reaching large effect level.

2. **Metric Dissociation**: ROUGE-L and quality judgment show dissociation—Qwen-Max has slight advantage in lexical overlap (d=-0.43) but significantly trails in content quality (d=+1.04), suggesting lexical-level metrics' limitations in creative writing evaluation.

3. **Structural Preservation**: GPT-5.2 demonstrates clear advantage in format convention compliance, with Structural Similarity 18.25 percentage points higher (0.93 vs 0.75, d=0.46).

4. **Quality Dimensions**: GPT-5.2 leads comprehensively across multiple sub-dimensions including Character Consistency (+22.38), Tone-Style (+26.69), and Format Match (+16.83).

5. **Stability**: GPT-5.2 shows higher generation stability, with better validity rate (98.7% vs 91.8%) and quality variance than Qwen-Max.

6. **Shared Challenges**: Both models perform poorly on Ending Closure, reflecting current LLMs' shared limitations in long-range narrative conclusion.

## 6.2 Future Work

1. **Human Evaluation Validation**: Introduce professional screenwriters for blind evaluation, establishing correlation analysis between LLM-as-Judge scores and human judgment to enhance evaluation system credibility.

2. **Dataset Expansion**: Add more film samples covering broader genres and eras, with stratified analysis to reveal differentiated model performance across script types.

3. **Multi-Model Comparison**: Extend evaluation scope to other mainstream models including Claude, Gemini, and GLM, establishing more comprehensive benchmarks.

4. **Fine-grained Tasks**: Design scene-level and dialogue-level generation and evaluation tasks for deeper analysis of model capabilities across script creation stages.

5. **Intervention Experiments**: Explore effects of different prompt strategies, temperature parameters, and context lengths on generation quality, providing parameter tuning guidelines for practical applications.

6. **Long-Range Narrative Improvement**: Address universally low Ending Closure scores by exploring

hierarchical generation, outline-guided strategies, and other approaches to improve long-range narrative planning capabilities.

# Appendix

## Appendix A: Dataset Film List (Partial)

| ID | Film Title | Year | Genre | First-Half Chars | Second-Half Chars |
|---|---|---|---|---|---|
| 01 | Detective Chinatown | 2015 | Comedy/Mystery | 21,322 | 17,078 |
| 02 | Farewell My Concubine | 1993 | Drama | 34,375 | 24,033 |
| 04 | Bronze Swallow Terrace | 2012 | Historical | 21,633 | 13,258 |
| 05 | Detective Chinatown 2 | 2018 | Comedy/Mystery | 18,222 | 14,326 |
| 06 | Happy Together | 1997 | Drama | 6,995 | 5,766 |
| 28 | Chungking Express | 1994 | Drama | 35,426 | 23,102 |
| 29 | Buddies in India | 2017 | Comedy | 37,422 | 34,453 |
| ... | ... | ... | ... | ... | ... |
| 62 | Eat Drink Man Woman | 1994 | Drama | 29,125 | 21,915 |

(Complete list contains 53 films)

## Appendix B: Format Profile Example

Using *Detective Chinatown* (ID=01) as example:

```
{
  "scene_header_style": "NONE",
  "dialogue_marker": "ROLE_COLON",
  "stage_direction_marker": "PAREN",
  "blankline_policy": "SINGLE_NEWLINE",
  "emphasis_style": "MARKDOWN_BOLD",
  "examples": {
    "scene_headers": [],
    "dialogues": [
      "Male Examiner: Qin Feng, why do you want to apply to the Criminal Police Academy",
      "Female Examiner: Is this question difficult to answer?"
```

```
      ]
    }
  }
```

## Appendix C: Scoring Dimension Definitions

| Dimension | Definition | Score Range |
|---|---|---|
| overall_similarity | Comprehensive similarity considering plot, character, style, format | 0-100 |
| plot_event_alignment | Alignment of key events, plot turns with actual second half | 0-100 |
| character_consistency | Consistency of character personality, motivation, relationships | 0-100 |
| tone_style_match | Match of tone, atmosphere, narrative style | 0-100 |
| format_match | Match of scene headings, dialogue format, stage directions, etc. | 0-100 |
| ending_closure | Narrative conclusion, thematic resonance, ending completeness | 0-100 |

## Appendix D: Statistical Methods

**Paired t-test Assumptions**:

- Normality of differences: Verified via Shapiro-Wilk test
- Sample independence: Each pair from same film, same sample_idx

**Effect Size Calculation**:

$$d = \frac{\bar{X}_A - \bar{X}_B}{SD_{\text{diff}}}$$

where $\bar{X}_A$ and $\bar{X}_B$ are the two group means, and $SD_{\text{diff}}$ is the standard deviation of differences.

**95% Confidence Interval**:

$$CI = (\bar{X}_A - \bar{X}_B) \pm 1.96 \times \frac{SD_{\text{diff}}}{\sqrt{n}}$$

---

### Acknowledgments

We thank OpenAI and Alibaba Cloud for API service support. We thank the DeepSeek team for developing the Reasoner model that provided high-quality evaluation capability for this study.

---

### Author Contributions

Y. Cao and Z. Yang contributed equally to this work, including conceptualization, methodology design, data collection, analysis, and manuscript writing. Y. Wang supervised the project and provided critical revision.

**Corresponding Author**

Ye Wang, School of Journalism and Communication, Wuhan University, Wuhan 430072, China.
Email: [wangye@whu.edu.cn](wangye@whu.edu.cn)

**Data Availability**

The dataset and code are available upon reasonable request to the corresponding author.

---

*Submission Date: January 21, 2026*